\documentclass[a4paper,twoside]{article}
\usepackage{epsfig}
\usepackage{subcaption}
\usepackage{calc}

\usepackage{amssymb}
\usepackage{amstext}
\usepackage{amsmath}
\usepackage{amsthm}
\usepackage{multicol}
\usepackage{pslatex}
\usepackage{apalike}
\usepackage[bottom]{footmisc}
\usepackage{SCITEPRESS}
\usepackage{amsmath}
\usepackage{bbding}
\usepackage{natbib}
\usepackage{breqn}
\usepackage{hyperref}

\usepackage{algorithm}
\usepackage{algorithmic}
\usepackage{todonotes}
\usepackage{bbm}
\usepackage{graphicx}
\usepackage{booktabs}

\begin{document}

\title{First Go, then Post-Explore: \\
the Benefits of Post-Exploration in Intrinsic Motivation}

\author{\authorname{Zhao Yang, Thomas M. Moerland, Mike Preuss and Aske Plaat}
\affiliation{Leiden University, the Netherlands}
\email{z.yang@liacs.leidenuniv.nl}
}


\abstract{Go-Explore achieved breakthrough performance on challenging reinforcement learning (RL) tasks with sparse rewards. The key insight of Go-Explore was that successful exploration requires an agent to first return to an interesting state (`Go'), and only then explore into unknown terrain (`Explore'). We refer to such exploration after a goal is reached as `post-exploration'. In this paper, we present a clear ablation study of post-exploration in a general intrinsically motivated goal exploration process (IMGEP) framework, that the Go-Explore paper did not show. We study the isolated potential of post-exploration, by turning it on and off within the same algorithm under both tabular and deep RL settings on both discrete navigation and continuous control tasks. Experiments on a range of MiniGrid and Mujoco environments show that post-exploration indeed helps IMGEP agents reach more diverse states and boosts their performance. In short, our work suggests that RL researchers should consider using post-exploration in IMGEP when possible since it is effective, method-agnostic, and easy to implement.}

\onecolumn \maketitle \normalsize \setcounter{footnote}{0} \vfill

\section{\uppercase{Introduction}} \label{introduction}
\begin{figure*}[t]
    \centering
    \includegraphics[scale=0.35]{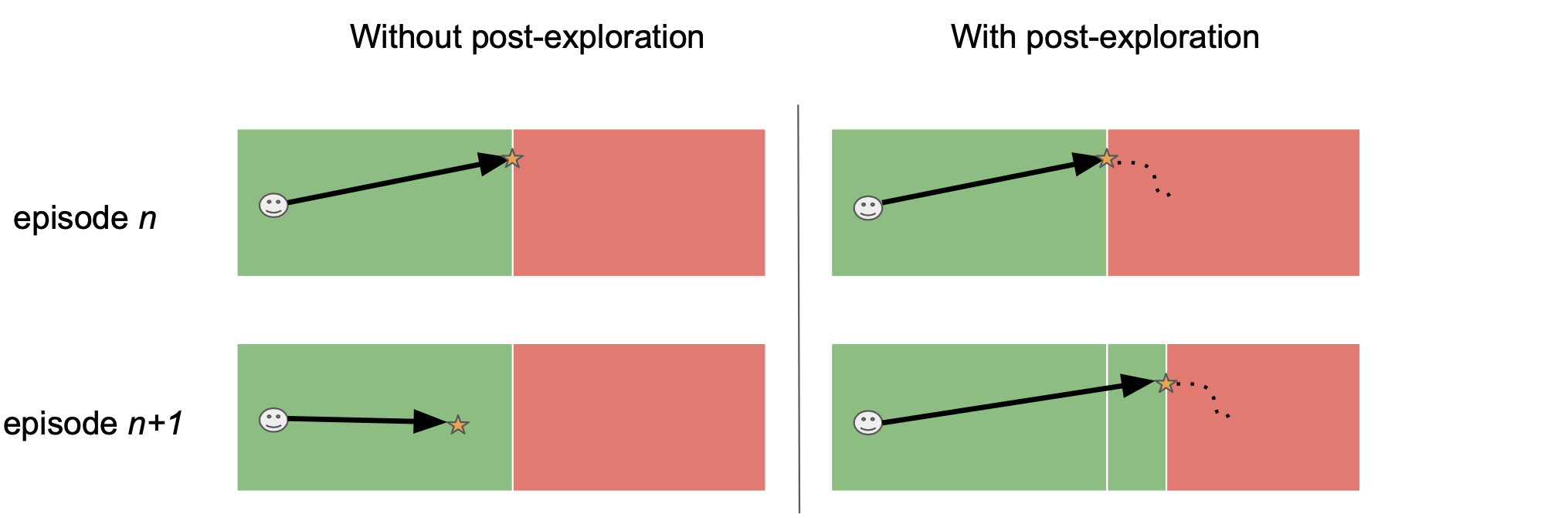}
    \caption{Conceptual illustration of post-exploration. Each box displays the entire state space, where green and red denote the (currently) explored and unexplored regions, respectively. {\bf Left}: Goal-based exploration without post-exploration. The top graph shows the agent reaching a current goal, after which the episode is terminated (or the next goal in the green area is chosen). Therefore, in the next episode (bottom) the agent will again explore within the known (green) region), often leaving the unknown (red) area untouched. {\bf Right}: Goal-based exploration with post-exploration. The top figure shows the agent reaching a particular goal, from which the agent now post-explores for several steps (dashed lines). Thereby, the known area (green) is pushed outwards. On the next episode, the agent may now also select a goal in the expanded region, gradually expanding its knowledge of the reachable state space.}
    \label{fig:intuition}
\end{figure*}

Go-Explore~\citep{ecoffet2021first} achieved breakthrough performance on challenging reinforcement learning (RL) tasks with sparse rewards, most notably achieving state-of-the-art, `super-human' performance on Montezuma's Revenge, a long-standing challenge in the field. The key insight behind Go-Explore is that proper exploration should be structured in two phases: an agent should first attempt to get back to a previously visited, interesting state (`Go'), and only then explore into new, unknown terrain (`Explore'). Thereby, the agent gradually expands its knowledge base, an approach that is visualized in Fig.\ref{fig:intuition}. We propose to call such exploration after the agent reached a goal {\it post-exploration} (to contrast it with standard exploration).

There are actually two variants of Go-Explore in the original paper: one in which we directly reset the agent to an interesting goal ({\it restore-based} Go-Explore), and one in which the agent has to act to get back to the proposed goal ({\it policy-based} Go-Explore). In this work, we focus on the latter approach, which is technically part of the literature on intrinsic motivation, in particular intrinsically motivated goal exploration processes (IMGEP)~\citep{colas2020intrinsically}. Note that post-exploration does not require any changes to the IMGEP framework itself, and can therefore be easily integrated into other existing work in this direction. 

While Go-Explore gave a strong indication of the potential of post-exploration, it did not structurally investigate the benefit of the approach. Go-Explore was compared to other baseline algorithms, but post-exploration itself was never switched on and off in the same algorithm. Thereby, the isolated performance gain of post-exploration remains unclear. 

Therefore, the present paper performs an ablation study of post-exploration in both tabular and deep RL settings on both discrete and continuous tasks. Experiments in a range of MiniGrid and Mujoco tasks show that post-exploration provides a strong isolated benefit over standard IMGEP algorithms. As a smaller contribution, we also cast Go-Explore into the IMGEP framework, and show how it can be combined with hindsight experience replay (HER) ~\citep{NIPS2017_453fadbd} to make more efficient use of the observed data. In short, our work presents a systematic study of post-exploration and shows its effectiveness on both discrete navigation tasks and continuous control tasks.

\section{\uppercase{Related Work}}
\label{related}
Go-Explore is a variant of intrinsically motivated goal exploration processes (IMGEP) (surveyed by~\citep{colas2020intrinsically}) which generally consists of three phases: 1) defining/learning a goal space, 2) sampling an interesting goal from the goal space, and 3) getting to the sampled goal based on knowledge from previous episodes. Regarding the first step, Go-Explore~\citep{ecoffet2021first} pre-defines goals as down-scaled states, and the goal space is adaptively extended by adding new encountered states. Other IMGEP approaches, like Goal-GAN~\citep{pmlr-v80-florensa18a} uses a generative adversarial network to learn a goal space of appropriate difficulty. As an alternative, AMIGo~\citep{campero2021learning} trains a teacher to propose goals for a student to reach. In this work, we largely follow the ideas of Go-Explore, and pre-define the goal space as the visited state space, since our research questions are directed at the effects of post-exploration.

After a goal space is determined, we need a sampling strategy to set the next goals. Generally, desired next goals should be novel/diverse enough or lie on the border of the agent's knowledge boundary. Example strategies to select next goals include novelty~\citep{ecoffet2021first,pitis2020maximum}, learning progress~\citep{colas2019curious,portelas2020teacher},  diversity~\citep{pong2019skew,warde-farley2018unsupervised}, and uniform sampling~\citep{eysenbach2018}. In this work, we use novelty (count-based) for sampling, since it has less tuneable hyper-parameters.

In the third IMGEP phase, we need to get back to the proposed goal. One variant of Go-Explore simply resets the agent to the desired goal, but this requires an environment that can be set to an arbitrary state. Instead, we follow the more generic `policy-based Go-Explore' approach, which uses a goal-conditioned policy network to get back to a goal. The concept of goal-conditioning, which was introduced by~\citep{schaul2015universal}, is also a common approach in IMGEP approaches. A well-known addition to goal-based RL approaches is hindsight experience replay~\citep{NIPS2017_453fadbd}, which allows the agent to make more efficient use of the data through counterfactual goal relabelling. Although Go-Explore did not use hindsight in their work, we do include it as an extension in our approach. 

After we manage to reach a goal, most IMGEP literature samples a new goal and either reset the episode or continue from the reached state. The main contribution of Go-Explore was the introduction of post-exploration, in which case we temporarily postpone the selection of a new goal (Tab.~\ref{tab:overview}). While post-exploration is a new phenomenon in reinforcement learning literature, it is a common feature of most planning algorithms, where we for example select a particular node on the frontier, go there first, and then unfold the search tree into unknown terrain. 

\begin{table*}[t]
    \centering
    \caption{Overview of moment of exploration in IMGEP papers. Most IMGEP approaches, like Goal-GAN~\citep{pmlr-v80-florensa18a}, CURIOUS~\citep{colas2019curious}, DIAYN~\citep{pong2019skew}, MEGA~\citep{pitis2020maximum} and AMIGo~\citep{campero2021learning}, only explore during goal reaching. Restore-based Go-Explore, which directly resets to a goal, only explore after the goal is reached. Our work, similar to policy-based Go-explore, explores both during goal-reaching and after goal reaching.}
    \begin{tabular}{c|c|c}
        \textbf{Approach} & \textbf{Exploration} & \textbf{Post-exploration} \\
        \hline
        IMGEP & \Checkmark & \XSolidBrush \\
        Restore-based Go-Explore & \XSolidBrush & \Checkmark \\
        Policy-based Go-Explore + this paper & \Checkmark & \Checkmark \\
    \end{tabular}

    \label{tab:overview}
\end{table*}

\section{\uppercase{Background}}
We adopt a Markov Decision Processes (MDPs) formulation defined as the tuple $M=(\mathcal{S}, \mathcal{A}, P, R, \gamma)$ ~\citep{sutton2018reinforcement}. Here, $\mathcal{S}$ is a set of states, $\mathcal{A}$ is a set of actions the agent can take, $P$ specifies the transition function of the environment, and $R$ is the reward function. At timestep $t$, the agent observes state $s_t \in \mathcal{S}$, selects an action $a_t \in \mathcal{A}$, after which the environment returns a next state $s_{t+1} \sim P(\cdot|s_t,a_t)$ and associated reward $r_t = R(s_t,a_t,s_{t+1})$. We act in the MDP according to a policy $\pi(a|s)$, which maps a state to a distribution over actions. When we unroll the policy and environment for $T$ steps, define the cumulative reward (return) of the resulting trace as  $\sum ^{T}_{k=0} \gamma^k \cdot r_{t+k+1}$, where $\gamma \in [0,1]$ denotes a discount factor. The goal in RL is to learn a policy that can maximize the expected return. 

Define the state-action value $Q^\pi(s,a)$ as the {\it expected} cumulative reward under some policy $\pi$ when we start from state $s$ and action $a$, i.e.,

\begin{equation}
    Q^\pi(s_t,a_t) = \mathbb{E}_{\pi,P} \Big[\sum ^{T}_{k=0} \gamma^k \cdot r_{t+k}|s_t=s,a_t=a \Big]
\end{equation} 

Our goal is to find the {\it optimal} state-action value function $Q^*(s,a)=max_{\pi} Q^\pi(s,a)$, from which we may at each state derive the optimal policy $\pi^*(s)=argmax_a Q^*(s,a)$. A well-known RL approach to solve this problem is {\it Q-learning} \citep{watkins1992q}. Tabular Q-learning maintains a table of Q-value estimates $\hat{Q}(s,a)$, collects transition tuples $(s_t,a_t,r_t,s_{t+1})$, and subsequently updates the tabular estimates according to 
\begin{dmath}
    \hat{Q}(s_t,a_t) \gets \hat{Q}(s_t,a_t) + \alpha \cdot[r_{t}+\gamma \cdot \max_{a} \hat{Q}(s_{t+1},a) - \hat{Q}(s_t,a_t)]
    \label{eq:q}
\end{dmath} where $\alpha \in [0,1]$ denotes a learning rate. Under a policy that is greedy in the limit with infinite exploration (GLIE) this algorithm converges on the optimal value function \citep{watkins1992q}.

In deep RL, state-action value function $Q(s,a)$ is approximated by a parameterized neural network, denoted by $Q_{\phi}(s,a)$. While dealing with continuous action space, taking $max$ operation over state-action values like in Eq.~\ref{eq:q} is intractable, we could learn a parameterized deterministic policy $\mu_{\theta}$ for the agent instead. A deterministic policy maps a state to a deterministic action rather than a distribution over all actions. Then we can approximate $\max_a Q_{\phi}(s,a)$ with $Q_{\phi}(s,\mu_\theta(s))$.

\begin{dmath}
    L(\phi, D) = \mathbb{E}_{(s_t,a_t,r_t,s_{t+1})\sim D} \Big(r_{t}+\gamma \cdot \hat{Q}(s_{t+1},\mu(s_{t+1})) - \hat{Q}(s_t,a_t)\Big )^2
    \label{eq:critic}
\end{dmath}

Deep deterministic policy gradient (DDPG)~\citep{DBLP:journals/corr/LillicrapHPHETS15} is an actor-critic method that parameterizes the deterministic policy (`actor') and the state-action value function (`critic') as neural networks. The critic is updated by minimizing the mean squared TD error $L(\phi, D)$ in Eq.~\ref{eq:critic}, where $D$ is a set of collected transitions. The actor is updated by solving the Eq.~\ref{eq:actor}.

\begin{equation}
    \max_{\theta} \mathbb{E}_{s_t\sim D} \Big[Q_{\phi}(s_t,\mu_{\theta}(s_t))\Big]
    \label{eq:actor}
\end{equation}

\section{\uppercase{Methods}}
We will first describe how we cast Go-Explore into the general IMGEP framework (Sec. \ref{sec_imgep}), and subsequently introduce how we do post-exploration (Sec. \ref{sec_post_exploration}). 

\begin{figure*}[t]
    \centering
    \includegraphics[scale=0.6]{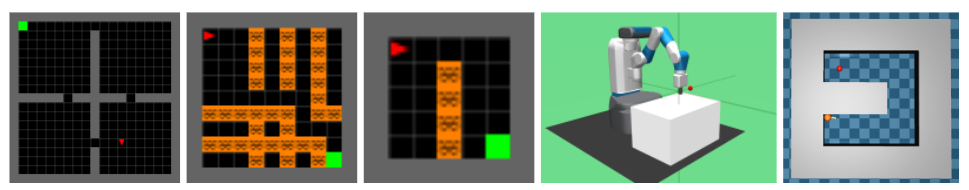}
    \caption{FourRooms, LavaCrossing, LavaGap, FetchReach, and PointUMaze. Room layouts of two lava environments are procedurally generated.}
    \label{fig:envs}
\end{figure*}
\subsection{IMGEP} \label{sec_imgep}
\begin{algorithm}[!htb]
   \caption{IMGEP with post-exploration and hindsight (\textcolor{red}{red part is post-exploration})}
   \label{alg:pe}
\begin{algorithmic}
   \STATE {\bfseries Initialize:} Goal-conditioned RL agent $RL$, environment \texttt{Env}, episode memory $M$, goal space $G$, goal sampling probability $p(g)$, goal-conditioned reward function $R_g(s,a,s')$.
   \WHILE{training budget left}
   \STATE $g \sim p(g)$ \hfill\COMMENT{Sample a goal $g$ from the goal space $G$ according to the goal sampling probability $p(g)$.}
   \STATE $s = \texttt{Env}$.reset() \hfill\COMMENT{Reset the environment.}
   \STATE Execute a roll-out using $RL$ and collect transitions $T$ along the way.
   \STATE Update $G$ and $p(g)$ using $T$. \hfill\COMMENT{Extend $G$ by adding new encountered states from $T$.}
   \STATE Store collected transitions $T$ to $M$.
   \IF{\textcolor{red}{$g$ is reached}}
    \STATE \textcolor{red}{Execute a fixed number of steps using the random policy and collect transitions $T_r$ along the way.}
    \STATE \textcolor{red}{Update $G$ and $p(g)$ using transitions $T_r$.}
   \STATE \textcolor{red}{Store collected transitions $T_r$ to $M$.}
   \ENDIF
   \STATE Sample a batch $B$ from $M$ and hindsight relabel according to the reward function $R_g$.
   \STATE Update RL agent $RL$ using batch $B$.
\ENDWHILE
\end{algorithmic}
\end{algorithm}

Our work is based on the IMGEP framework shown in Alg~\ref{alg:pe}. Since this paper attempts to study the fundamental benefit of post-exploration, we try to simplify the problem setting as much as possible, to avoid interference with other issues (such as goal space representation learning, goal sampling strategy design, reward function design, etc). We, therefore, define the goal space as the set of states we have observed so far. For continuous tasks, we discretize the whole state space into bins, so the goal space for continuous control tasks is the set of bins we have observed so far. Firstly, a bin will be sampled, then we further uniformly sample a goal from that bin. The goal space $G$ is initialized by executing a random policy for one episode, while new states/bins in future episodes augment the set. 

For goal sampling, we use a count-based novelty sampling strategy which aims to sample goals that are less visited more often and more visited less often from the goal space. We therefore track the number of times $n(g)$ we visited a particular goal $g$, and specify the probability of sampling it $p(g)$ as 
\begin{equation}\label{GoalSampling}
    p(g)_{g\in G} = \frac{(\frac{1}{n(g)})^\tau}{\sum\limits_{g\in G}(\frac{1}{n(g)})^\tau}.
\end{equation}
Here, $\tau$ is a temperature parameter that allows us to scale goal sampling. When $\tau$ is 0, goals will be sampled uniformly from the goal space, while larger $\tau$ will emphasize less visited goals more.

Note that we could also use other methods, both for defining the goal space and for sampling new goals (for example based on curiosity or learning progress). 

To get (back) to a selected goal, we train a goal-conditioned policy, for tabular settings we use Q-learning, and for deep RL settings we use DDPG. Agents are trained on the goal-conditioned reward function $R_g$, which is a one-hot indicator that fires when the agent manages to reach the specified goal: 

\begin{equation}
    R_g(s,a,s') =  \mathbbm{1}_{s'=g}.
\end{equation} 

Note that different goal-conditioned rewards functions, and different update methods are of course possible as well. 

\begin{figure*}[t]
   \begin{minipage}{0.32\textwidth}
     \centering
     \includegraphics[width=.9\linewidth]{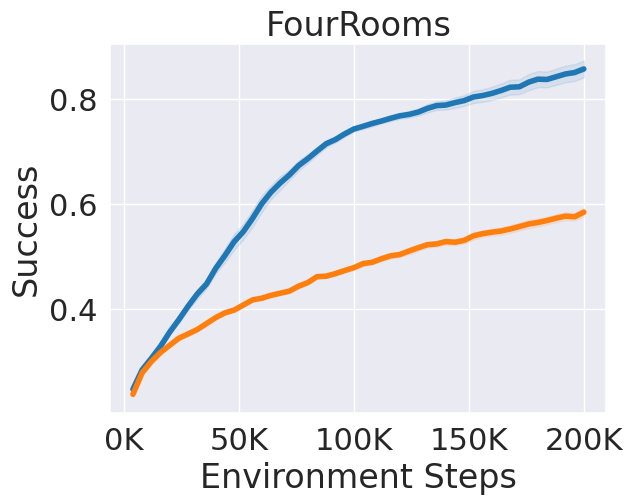}
   \end{minipage}
   \begin{minipage}{0.32\textwidth}
     \centering
     \includegraphics[width=.9\linewidth]{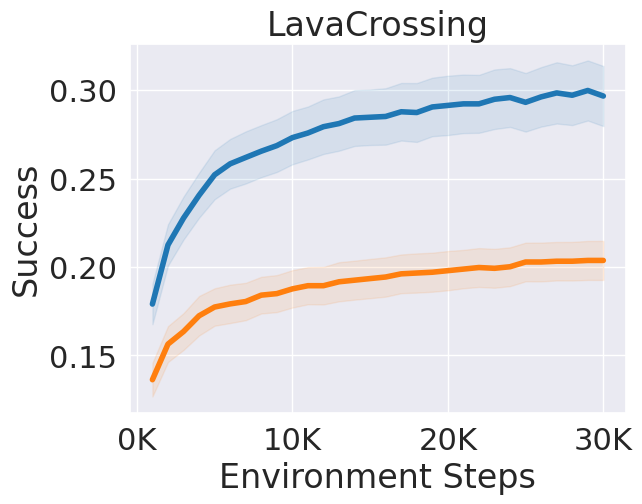}
   \end{minipage}
   \begin{minipage}{0.32\textwidth}
     \centering
     \includegraphics[width=.9\linewidth]{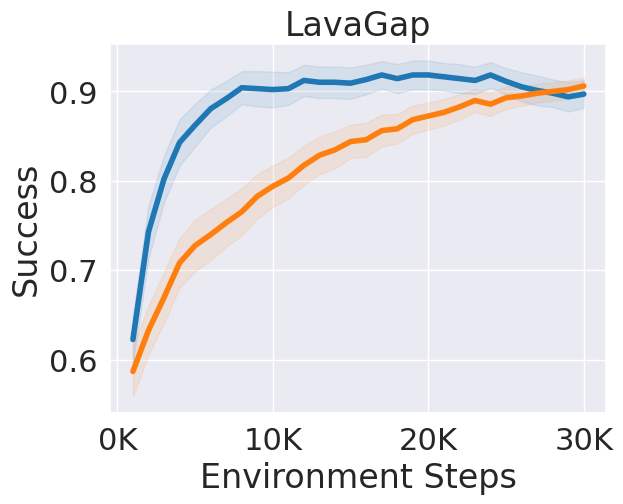}
   \end{minipage}
   \caption{IMGEP agents with (blue) and without (orange) post-exploration. {\bf Left}: Performance on FourRooms, with results averaging over three different values of $\epsilon$ ($0,0.1,0.3$). {\bf Middle}: Performance on LavaCrossing, averaged over 10 different environment seeds. {\bf Right}: Performance in LavaGap, averaged over 10 different environment seeds. Overall, agents with post-exploration outperform agents without post-exploration.}
   \label{fig: pe_grid}
\end{figure*}

\subsection{Post-exploration} \label{sec_post_exploration}
\label{PE}
The general concept of post-exploration was already introduced in Figure~\ref{fig:intuition}. The main benefit of this approach is that it increases our chance to step into new, unknown terrain. Based on this intuition, we will now discuss when and how to post-explore, as well as how we utilize data collected during the post-exploration.

As is directly visible from Figure~\ref{fig:intuition}, post-exploration is aiming to expand the agent's knowledge boundary further when necessary. Since the less visited goals are sampled more often, when a selected goal is reached by the agent, that means the agent's knowledge boundary (the ability to reach goals) is already near the selected goal, then we decide to expand the boundary by doing post-exploration. In this work, we simply let the agent post-explore every time the selected goal is reached. More constraints can be applied here, such as only doing post-exploration when the reached goal is lying on the border of the unknown area or it is the bottleneck of a graph of the visited states, etc. We leave the design of smarter strategies for future work.

We assume that post-exploration could lead the agent to step into new naive areas, thus it is important to design an effective way to do post-exploration. In Go-Explore~\citep{ecoffet2021first}, they mentioned random post-exploration is a simple and effective way. So in this work, we also use the random policy during the post-exploration. The agent will take a fixed number of random actions for post-exploration.

The agent in principle only observes a non-zero reward when it successfully reaches the goal, which may not happen at every attempt. We may improve sample efficiency (make more efficient use of each observed trajectory) through {\it hindsight experience relabelling}(HER)~\citep{NIPS2017_453fadbd}. With hindsight, we imagine states in our observed trajectory were the goals we actually attempted to reach, which allows us to extract additional information from them (`if my goal had been to get to this state, then the previous action sequence would have been a successful attempt'). For tabular settings, we choose to always relabel 50\% of the entire trajectory (goal reaching plus post-exploration),  i.e., half of the states in each trajectory are imagined as if they were the goal of that episode. We choose to always relabel the full post-exploration part, because it likely contains the most interesting information, and randomly sample the remaining relabelling budget from the part of the episode before the goal was reached. For deep RL settings, we use the `future' relabel strategy from the original HER work~\citep{NIPS2017_453fadbd}.

\section{\uppercase{Experiments}}
\label{sec:experiments}

\subsection{Experimental Setup}
We test our work on three MiniGrid environments~\citep{gym_minigrid} (tabular settings), and two Mujoco environments~\citep{pointmaze,1802.09464} (with using the function approximation), visualized in Fig.~\ref{fig:envs}. Results on two lava environments are averaged over 10 seeds, i.e., 10 independently drawn instances of the task. For evaluation, not like in Go-Explore where the agent needs to reach a specific final goal (the state with the highest score), we instead test the ability of the agent to reach every possible state in the whole state space. This checks to what extent the goal-conditioned agent is able to reach a given goal, when we execute the greedy policy (turn exploration off). All our results report the total number of environment steps on the x-axis. Therefore, since an episode with post-exploration takes longer, it will also contribute more steps to the x-axis (i.e., we report performance against the total number of unique environment calls). Curves display mean performance over time, including the standard error over 5 repetitions for each experiment (10 repetitions for DRL experiments).

\subsection{Results}
Fig.~\ref{fig: pe_grid} compares an IMGEP agent with post-exploration to an IMGEP agent without exploration in the three MiniGrid environments. On all three environments, the agent with post-exploration outperforms the baseline agent significantly (on its average ability to reach an arbitrary goal in the state space). 

\begin{figure*}[!t]
    \centering
    \includegraphics[scale=0.35]{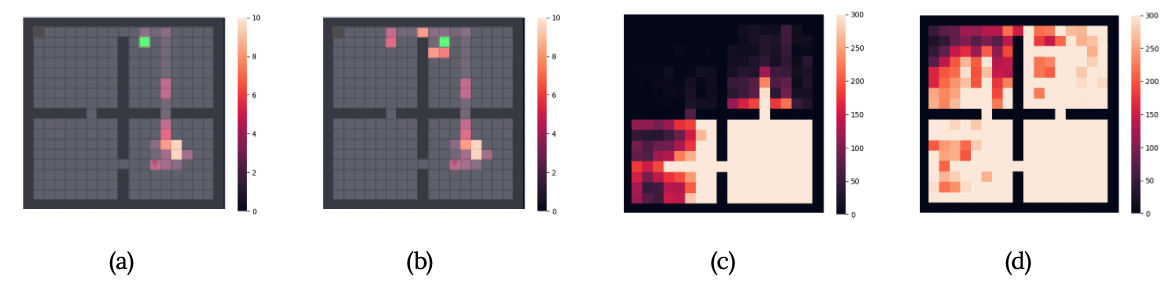}
    \caption{Comparison of characteristic traces (a,b) and coverage (c,d) for agents without post-exploration (a,c) and with post-exploration (b,d). Colour bars indicate the number of visitations, green square indicates the selected goal in a particular episode. (a): Standard exploration towards a goal without post-exploration. (b): With post-exploration, the agent manages to reach the next room. (c): Coverage after 200k training steps without post-exploration. (d): Coverage after 200k training steps with post-exploration. The boundary of the coverage with post-exploration clearly lies further ahead.}
    \label{fig:peNcover}
\end{figure*}

A graphical illustration of the effect of post-exploration is provided in Fig.~\ref{fig:peNcover}. Part a shows a characteristic trace for an agent attempting to reach the green square {\it without} post-exploration, while b shows a trace for an agent {\it with} post-exploration. In a, we see that the agent is able to reach the goal square, but exploration subsequently stops, and the agent did not learn anything new. In part b, the agent with post-exploration subsequently manages to enter the next new room. This graph, therefore, gives a practical illustration of our intuition from Fig.~\ref{fig:intuition}.

To further illustrate this effect, in Fig.~\ref{fig:peNcover}, c and d show a visitation heat map for the agent after 200k training steps, without post-exploration (c) and with post-exploration (d). The agent with post-exploration (c) has primarily explored goals around the start region (in the bottom-right chamber). The two rooms next to the starting room have also been visited, but the agent barely managed to get into the top-left room. In contrast, the agent with post-exploration (d) has extensively visited the first three rooms, while the coverage boundary has also been pushed into the final room already. This effect translates to the increased goal-reaching performance of post-exploration visible in Fig.~\ref{fig: pe_grid}. We think such a principle holds in other environments as well.

\begin{figure}[!t]
   \begin{minipage}{0.23\textwidth}
     \centering
     \includegraphics[width=.98\linewidth]{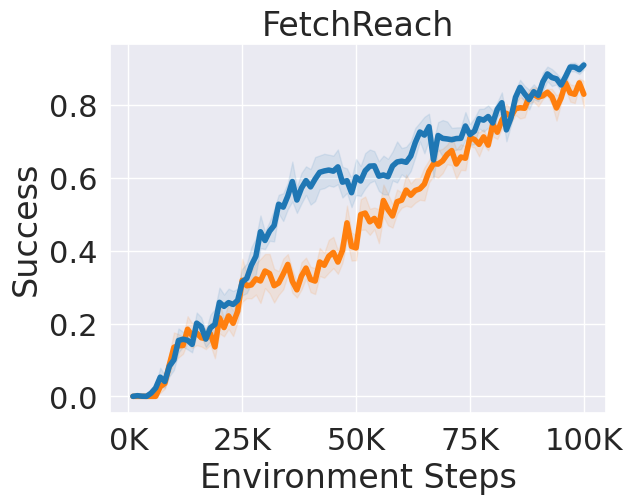}
   \end{minipage}
   \begin{minipage}{0.23\textwidth}
     \centering
     \includegraphics[width=.98\linewidth]{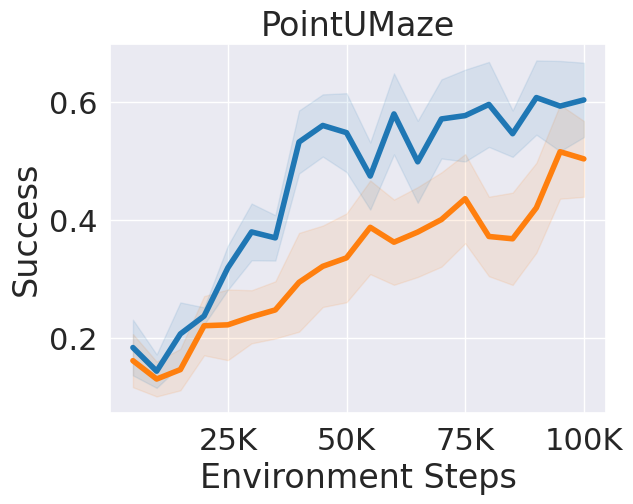}
   \end{minipage}
   \caption{IMGEP agents with (blue) and without (orange) post-exploration. {\bf Left}: Performance on FetchReach, with results averaging over 10 different random seeds. {\bf Right}: Performance on PointUMaze, averaged over 10 different random seeds. Overall, agents with post-exploration outperform agents without post-exploration.}
   \label{fig: pe_mujoco}
\end{figure}

Comparisons between agents with post-exploration and ones without on Mujoco environments are shown in Fig.~\ref{fig: pe_mujoco} and we use the DDPG agent with HER as the baseline (orange lines in Fig.~\ref{fig: pe_mujoco}) to compare with. With post-exploration, agents (blue lines in Fig.~\ref{fig: pe_mujoco}) outperform baseline agents in both FetchReach and PointUMaze environments. We observed that at the beginning and the end, two agents (with post-exploration and without) tend to have similar performances. We interpret the phenomenon as follows: the agent is not able to reach any selected goals in the beginning so there are no post-exploration and no benefits; then gradually the agent starts being able to reach selected goals and post-exploration comes in thus further improving the performance; in the end, when the whole state space is been entirely discovered post-exploration will not add benefits anymore. We think this phenomenon happens in environments where the baseline agent (without post-exploration) can also fully discover the whole state space over time, however, it would less happen in environments where there are many bottlenecks so that the baseline agent can hardly fully discover the entire state space.

\begin{figure}[htb]
   \begin{minipage}{0.23\textwidth}
     \centering
     \includegraphics[width=.98\linewidth]{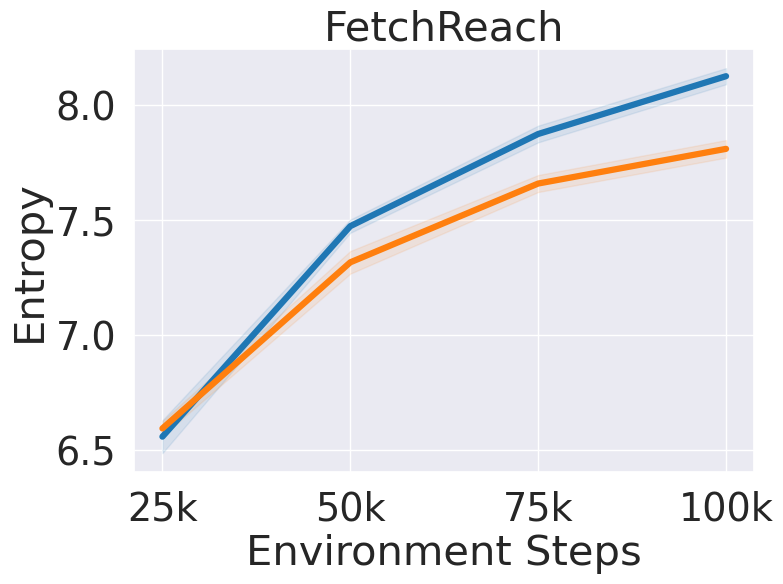}
   \end{minipage}
   \begin{minipage}{0.23\textwidth}
     \centering
     \includegraphics[width=.98\linewidth]{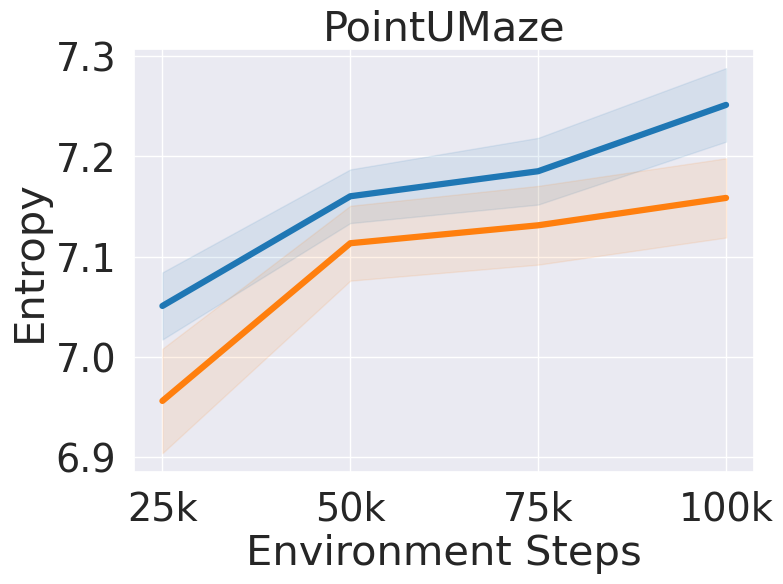}
   \end{minipage}
   \caption{The entropy of the visitation on the state space for the agent with (blue) and without (orange) post-exploration. {\bf Left}: Entropy on FetchReach, with results averaging over 10 different random seeds. {\bf Right}: Entropy on PointUMaze, averaged over 10 different random seeds. Adding post-exploration can increase the entropy of the agent's visitation on the state space.}
   \label{fig: entropy}
\end{figure}

By adding post-exploration, the agent is more likely to step into new unknown areas and thus can further improve the state coverage. Ideally, we want the agent's visitation of the state space to be as uniform as possible. More specifically, the higher entropy indicates that the visitation of the whole state space is more uniform. The entropy is computed using Eq.~\ref{eq:entropy}.
\begin{equation}
    -\sum\limits_{i=1}^Np(x_i)log(p(x_i))
    \label{eq:entropy}
\end{equation} where the whole state space is discretized into $N$ bins and $p(x_i)=n(x_i)/\sum\limits_{i=0}^Nn(x_i)$, $n(x_i)$ is the number of visitations of the $i^{th}$ bin.
Fig.~\ref{fig: entropy} shows the entropy on agents' visitation of the state space during the training. With post-exploration (blue lines), the entropy always lies above the one without (orange lines). Therefore, the principle (Fig.~\ref{fig:peNcover}) that we saw in the discrete navigation tasks is still true in continuous control tasks.

\section{\uppercase{Conclusion and Future Work}}
An intrinsically motivated agent not only needs to set interesting goals and be able to reach them but should also decide whether to continue exploration from the reached goal (`post-exploration'). In this work, we systematically investigated the benefit of post-exploration in the general IMGEP framework under different RL settings and tasks. Experiments in several MiniGrid and Mujoco environments show that post-exploration is not only beneficial in navigation tasks under tabular settings but also can be scaled up to more complex control tasks with neural networks involved. According to our results, agents with post-exploration gradually push the boundaries of their known region outwards, which allows them to reach a greater diversity of goals. Moreover, we realize that `post-exploration' is a very general idea and is easy to be plugged into any IMGEP method. Researchers should put more attention to this idea and consider using it when possible.

The current paper studied post-exploration in a simplified IMGEP framework, to better understand its basic properties. In the future, it would be interesting to plug post-exploration into other existing IMGEP methods directly to show its benefits. Moreover, our current implementation uses random post-exploration, which turned out to already work reasonably well. So an interesting direction for future work is to post-explore in a smarter way, like using macro actions or options. For example, in tasks where we need to control a more complex agent such as an ant or a humanoid robot, then random actions will never help the agent stand up properly and it is even harder to lead the agent step into new areas. Another promising future direction is to investigate the `adaptive' post-exploration. Intuitively, post-exploration will be most likely useful when it starts from a state that is new or important enough and the agent should post-explore more if the reached area is more naive, etc. In short, the agent should adaptively decide when and for how long to post-explore based on its own knowledge boundary or properties of reached goals. Altogether, post-exploration seems a promising direction for future RL exploration research.


\vskip 0.2in
\bibliographystyle{apalike}
{\small
\bibliography{example}}

\begin{thebibliography}{}

\bibitem[Andrychowicz et~al., 2017]{NIPS2017_453fadbd}
Andrychowicz, M., Wolski, F., Ray, A., Schneider, J., Fong, R., Welinder, P.,
  McGrew, B., Tobin, J., Pieter~Abbeel, O., and Zaremba, W. (2017).
\newblock Hindsight experience replay.
\newblock In Guyon, I., Luxburg, U.~V., Bengio, S., Wallach, H., Fergus, R.,
  Vishwanathan, S., and Garnett, R., editors, {\em Advances in Neural
  Information Processing Systems}, volume~30. Curran Associates, Inc.

\bibitem[Campero et~al., 2021]{campero2021learning}
Campero, A., Raileanu, R., Kuttler, H., Tenenbaum, J.~B., Rockt{\"a}schel, T.,
  and Grefenstette, E. (2021).
\newblock Learning with amigo: Adversarially motivated intrinsic goals.
\newblock In {\em International Conference on Learning Representations}.

\bibitem[Chevalier-Boisvert et~al., 2018]{gym_minigrid}
Chevalier-Boisvert, M., Willems, L., and Pal, S. (2018).
\newblock Minimalistic gridworld environment for openai gym.
\newblock \url{https://github.com/maximecb/gym-minigrid}.

\bibitem[Colas et~al., 2019]{colas2019curious}
Colas, C., Fournier, P., Chetouani, M., Sigaud, O., and Oudeyer, P.-Y. (2019).
\newblock Curious: intrinsically motivated modular multi-goal reinforcement
  learning.
\newblock In {\em International conference on machine learning}, pages
  1331--1340. PMLR.

\bibitem[Colas et~al., 2020]{colas2020intrinsically}
Colas, C., Karch, T., Sigaud, O., and Oudeyer, P.-Y. (2020).
\newblock Intrinsically motivated goal-conditioned reinforcement learning: a
  short survey.
\newblock {\em arXiv preprint arXiv:2012.09830}.

\bibitem[Ecoffet et~al., 2021]{ecoffet2021first}
Ecoffet, A., Huizinga, J., Lehman, J., Stanley, K.~O., and Clune, J. (2021).
\newblock First return, then explore.
\newblock {\em Nature}, 590(7847):580--586.

\bibitem[Eysenbach et~al., 2018]{eysenbach2018}
Eysenbach, B., Gupta, A., Ibarz, J., and Levine, S. (2018).
\newblock Diversity is all you need: Learning diverse skills without a reward
  function.

\bibitem[Florensa et~al., 2018]{pmlr-v80-florensa18a}
Florensa, C., Held, D., Geng, X., and Abbeel, P. (2018).
\newblock Automatic goal generation for reinforcement learning agents.
\newblock In Dy, J. and Krause, A., editors, {\em Proceedings of the 35th
  International Conference on Machine Learning}, volume~80 of {\em Proceedings
  of Machine Learning Research}, pages 1515--1528. PMLR.

\bibitem[kngwyu, 2021]{pointmaze}
kngwyu (2021).
\newblock mujoco-maze.
\newblock \url{https://github.com/kngwyu/mujoco-maze}.

\bibitem[Lillicrap et~al., 2016]{DBLP:journals/corr/LillicrapHPHETS15}
Lillicrap, T.~P., Hunt, J.~J., Pritzel, A., Heess, N., Erez, T., Tassa, Y.,
  Silver, D., and Wierstra, D. (2016).
\newblock Continuous control with deep reinforcement learning.
\newblock In Bengio, Y. and LeCun, Y., editors, {\em 4th International
  Conference on Learning Representations, {ICLR} 2016, San Juan, Puerto Rico,
  May 2-4, 2016, Conference Track Proceedings}.

\bibitem[Pitis et~al., 2020]{pitis2020maximum}
Pitis, S., Chan, H., Zhao, S., Stadie, B., and Ba, J. (2020).
\newblock Maximum entropy gain exploration for long horizon multi-goal
  reinforcement learning.
\newblock In {\em International Conference on Machine Learning}, pages
  7750--7761. PMLR.

\bibitem[Plappert et~al., 2018]{1802.09464}
Plappert, M., Andrychowicz, M., Ray, A., McGrew, B., Baker, B., Powell, G.,
  Schneider, J., Tobin, J., Chociej, M., Welinder, P., Kumar, V., and Zaremba,
  W. (2018).
\newblock Multi-goal reinforcement learning: Challenging robotics environments
  and request for research.

\bibitem[Pong et~al., 2019]{pong2019skew}
Pong, V.~H., Dalal, M., Lin, S., Nair, A., Bahl, S., and Levine, S. (2019).
\newblock Skew-fit: State-covering self-supervised reinforcement learning.
\newblock {\em arXiv preprint arXiv:1903.03698}.

\bibitem[Portelas et~al., 2020]{portelas2020teacher}
Portelas, R., Colas, C., Hofmann, K., and Oudeyer, P.-Y. (2020).
\newblock Teacher algorithms for curriculum learning of deep rl in continuously
  parameterized environments.
\newblock In {\em Conference on Robot Learning}, pages 835--853. PMLR.

\bibitem[Schaul et~al., 2015]{schaul2015universal}
Schaul, T., Horgan, D., Gregor, K., and Silver, D. (2015).
\newblock Universal value function approximators.
\newblock In {\em International conference on machine learning}, pages
  1312--1320. PMLR.

\bibitem[Sutton and Barto, 2018]{sutton2018reinforcement}
Sutton, R.~S. and Barto, A.~G. (2018).
\newblock {\em Reinforcement learning: An introduction}.
\newblock MIT press.

\bibitem[Warde-Farley et~al., 2019]{warde-farley2018unsupervised}
Warde-Farley, D., de~Wiele, T.~V., Kulkarni, T., Ionescu, C., Hansen, S., and
  Mnih, V. (2019).
\newblock Unsupervised control through non-parametric discriminative rewards.
\newblock In {\em International Conference on Learning Representations}.

\bibitem[Watkins and Dayan, 1992]{watkins1992q}
Watkins, C.~J. and Dayan, P. (1992).
\newblock Q-learning.
\newblock {\em Machine learning}, 8(3):279--292.

\end{thebibliography}


\clearpage
\appendix
\section{Hyper-parameter Settings}
\subsection{Tabular Q-learning}
All hyper-parameters we used for tabular Q-learning in this work are shown in Tab.~\ref{tab:para_Q}. Learning rate $\alpha$ and discount factor $\gamma$ are for Q-learning update (Eq.~\ref{eq:q}). $\tau$ is the temperature for goal sampling. (Eq.~\ref{GoalSampling}). $\epsilon$ is the exploration factor in $\epsilon$-greedy policy. We average results over 3 different values of $\epsilon$. $p_{pe}$ is the percentage of the whole trajectory that the agent will post-explore. $p_{pe}=0.5$ means the agent will post-explore $0.5 * l$ steps after the given goal is reached, $l$ is the length of the trajectory the agent takes to reach the given goal.
\begin{table}[!htb]
    \centering
    \begin{tabular}{c|c}
         \textbf{Parameters} & \textbf{Values} \\
         \hline
        learning rate & 0.1 \\
        discount factor & 0.99 \\
        $\tau$ & 0\\
        $\epsilon$ & 0, 0.1, 0.3 \\
        $p_{pe}$ & 0.5 \\
    \end{tabular}
    \caption{All hyper-parameters we used for tabular Q-learning experiments.}
    \label{tab:para_Q}
\end{table}

\subsection{DDPG}
All hyper-parameters we used for DDPG agent are shown in Tab.~\ref{tab:para_DDPG}. The learning rate $\alpha$ is for both the policy and the critic update. $\tau$ is the temperature for goal sampling. (Eq.~\ref{GoalSampling}). Random $\epsilon$ is the probability of taking a random action and noise $\epsilon$ is the probability of adding noise to selected actions. $n_{pe}$ is the number of steps that the agent will post-explore. $n_{bins}$ is the number of bins we discretize the whole goal space on each dimension. Batch size is the size of the batch sampled for training every time. Replay $k$ is the portion of data we will relabel. If $k=4$, that means we will relabel $4/5$ of the whole sampled training data. The replay strategy we used here is `future'.\footnote{For the remaining parameters of DDPG agents, we use the default settings from \url{https://github.com/TianhongDai/hindsight-experience-replay}.}
\begin{table}[!htb]
    \centering
    \begin{tabular}{c|c|c}
         \textbf{Parameters} & \textbf{FetchReach} & \textbf{PointUMaze}\\
         \hline
        learning rate & 0.001 & 0.001 \\
        discount factor & 0.98 & 0.98 \\
        $\tau$ & 0 & 0.01 \\
        random $\epsilon$ & 0.01 & 0.1 \\
        noise $\epsilon$ & 0.01 & 0.1 \\
        $n_{pe}$ & 30 & 50 \\
        $n_{bins}$ & 20 & 100 \\
        batch size & 16 & 2 \\
        replay $k$ & 4 & 4 \\
        replay strategy & future & future \\
    \end{tabular}
    \caption{All hyper-parameters we used for DDPG agents.}
    \label{tab:para_DDPG}
\end{table}

\end{document}